\documentclass{article}

\usepackage[final, nonatbib]{neurips_2023}

\usepackage[utf8]{inputenc} 
\usepackage[T1]{fontenc}    
\usepackage{hyperref}       
\usepackage{url}            
\usepackage{booktabs}       
\usepackage{amsfonts}       
\usepackage{nicefrac}       
\usepackage{microtype}      
\usepackage{xcolor}         

\usepackage{graphicx}
\usepackage{algorithm}

\usepackage{algcompatible}
\usepackage{algpseudocode}

\title{SPEED: Speculative Pipelined Execution for
Efficient Decoding}

\author{%
  \hspace{-1mm}Coleman Hooper\enskip\enskip\enskip
  Sehoon Kim \enskip\enskip\enskip
  Hiva Mohammadzadeh \enskip\enskip\enskip
  Hasan Genc\\
 \vspace{2mm}\textbf{Kurt Keutzer \enskip\enskip\enskip Amir Gholami \enskip\enskip\enskip Yakun Sophia Shao}\\
 University of California, Berkeley  \\
  \hspace{-3mm}\texttt{\small{\{chooper, sehoonkim, hiva, hngenc, keutzer, amirgh, ysshao\}@berkeley.edu}}\\
}

\newcommand{\OURS}{SPEED}

\usepackage{xcolor}

\begin{document}

\maketitle

\begin{abstract}

Generative Large Language Models (LLMs) based on the Transformer architecture have recently emerged as a dominant foundation model for a wide range of Natural Language Processing tasks. 
Nevertheless, their application in real-time scenarios has been highly restricted due to the significant inference latency associated with these models. 
This is particularly pronounced due to the autoregressive nature of generative LLM inference, where tokens are generated sequentially since each token depends on all previous output tokens. 
It is therefore challenging to achieve any token-level parallelism, making inference extremely memory-bound. 
In this work, we propose \OURS, which improves inference efficiency by speculatively executing multiple future tokens in parallel with the current token using predicted values based on early-layer hidden states. 
For Transformer decoders which employ parameter sharing, the memory operations for the tokens executing in parallel can be amortized, which allows us to accelerate generative LLM inference. 
We demonstrate the efficiency of our method in terms of latency reduction relative to model accuracy and demonstrate how speculation allows for training deeper decoders with parameter sharing with minimal runtime overhead.

\end{abstract}
\section{Introduction}
The Transformer neural network architecture has recently revolutionized NLP, providing massive accuracy gains across a range of tasks \cite{devlin2019bert,brown2020language}. 
In particular, there has been growing interest in applying Transformers decoders for generative tasks \cite{gpt2,t5}. 
Unlike Transformer encoders which can process an entire input sequence in parallel, Transformer decoders must be applied autoregressively at inference time as each input token depends on the output classification for the previous token. 
This means that they exhibit low arithmetic intensity and are typically memory bandwidth-bound \cite{full-stack-optimization, park2020optimus}. 
For small batch sizes (as is typical for edge deployment scenarios \cite{schuster2022confident}), it is extremely difficult to achieve any parallelism. In order to accelerate memory bandwidth-bound decoder inference, we must reduce the number of memory operations required. 

In this work, we aim to reduce the latency of memory bandwidth-bound decoder inference by employing \textit{speculative execution} in order to process tokens at different positions in the sequence \textit{in parallel}. 
When employing speculative execution, the forward passes for future tokens are started using speculative output values from earlier tokens. 
By starting future tokens, we can process them \textit{in parallel} with finishing the forward passes for earlier tokens.
If a prediction is later found to be wrong, we must invalidate all future inferences that were started based on the speculative output value from the incorrect prediction. 
By still following all iterations through to completion, we can ensure that full model accuracy is maintained.

\begin{figure*}[t!]
  \centering
  \includegraphics[width=\columnwidth]{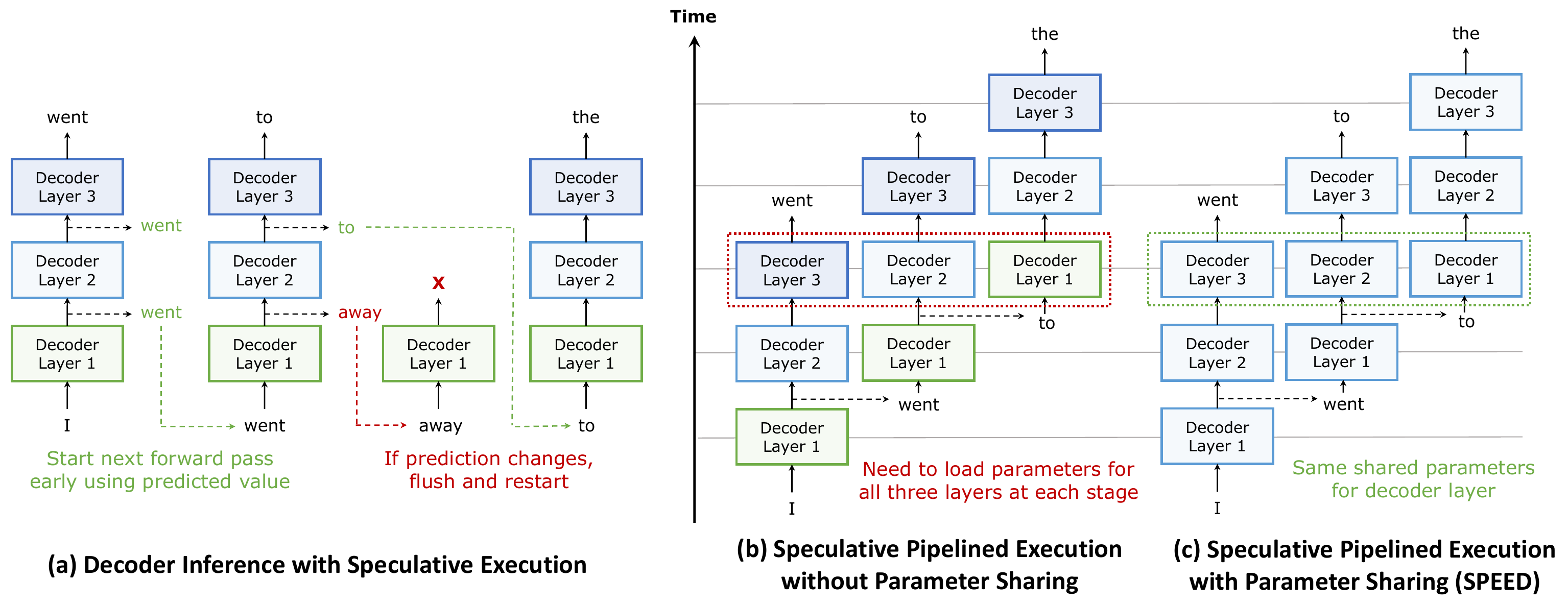}
  \caption{
  Outline of our methodology for speculative pipelined execution with parameter sharing. Diagram (a) shows how speculative values can be used to start later tokens, and how any incorrect predictions can later be corrected. Diagram (b) shows how this type of speculative execution allows us to pipeline inference, thereby achieving parallelism across the sequence length dimension. However, in a standard decoder, this doesn't help reduce memory operations since we would now need to load different decoder layers for different tokens in the sequence. Diagram (c) shows how in networks with parameter sharing, Speculative Pipelined Execution amortizes memory operations across the sequence length dimension, thereby allowing for Efficient Decoding (SPEED).}
      \label{fig:speculative-execution}
\end{figure*}

On its own, speculative execution would not lead to performance benefits within a single network. 
As shown in (b) in Figure \ref{fig:speculative-execution}, different tokens in the sequence would need to be processed by different layers in the network at the same time, meaning that the number of memory operations required for performing inference would not be reduced (even assuming perfect prediction). 
Additionally, to support inference on low-resource edge devices, it is crucial to reduce the model's memory footprint. \textit{Parameter sharing} is a common method for model compression in Transformer networks \cite{lan2020albert,reid2021subformer}. 
However, although it reduces the size of the network, parameter sharing doesn't typically provide significant speedup as the standard computation must still be performed for all layers in the network.
Even if inference is memory-bound, parameter sharing only reduces the number of memory operations required if the entire model fits in local cache memory, which is restrictive and hardware-dependent.

However, in a network which employs parameter sharing, speculative execution allows us to amortize the memory operations required for the weight matrices across different tokens in the sequence.
By employing speculative execution in networks with parameter sharing, we can \textit{pipeline} inference, thereby reducing memory operations.
Each pipeline stage corresponds to passing several tokens at different positions through the same set of linear layers (since the parameters for these linear layers are shared across decoder blocks).
Our speculative execution approach therefore allows us to accelerate decoder inference with networks that employ parameter sharing as a model compression method.
We believe that our speculative execution approach can make parameter sharing an advantageous model compression strategy for \textit{both} shrinking the static model size and for accelerating inference. 

\section{Method}
\label{sec:method}

\begin{figure*}[t!]
  \centering
  \includegraphics[width=0.85\columnwidth]{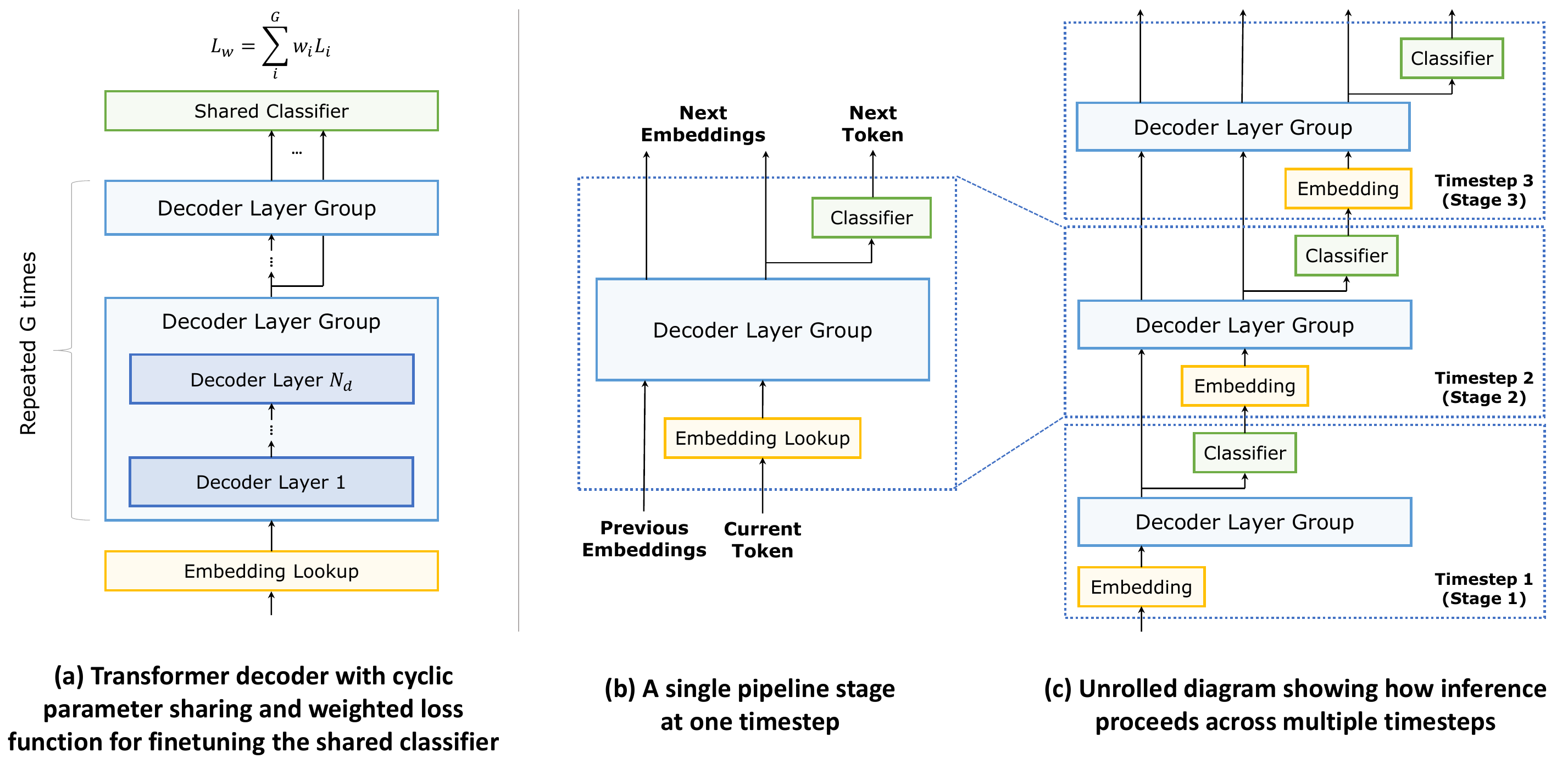}
  \caption{
  Diagram demonstrating how our speculative pipelined execution approach is implemented.  Diagram (a) shows how parameters are shared cyclicly across groups of decoder layers (where $N_d$ is the number of unique decoder layers shared $G$ times and $N=G*N_d$ is the total number of layers), and how the losses from the classifications at each layer are incorporated into a shared loss function during training. Note that we have omitted the final layer normalization layer prior to the shared classifier for simplicity. Diagram (b) shows a single pipeline stage in our inference process. Diagram (c) shows the progression of several pipeline stages in sequence.}
      \label{fig:implementation-overview}
\end{figure*}

\subsection{Parameter Sharing}
The parameter sharing scheme in this work corresponds to the ``CYCLE" configurations from \cite{takase2022lessons}, meaning that if a group of two decoder layers is shared three times, a forward pass consists of alternating between going through layer 1 and layer 2 three times.
Our cyclic parameter sharing scheme is outlined graphically in part (a) of Figure \ref{fig:implementation-overview}.
During fine-tuning, we incorporate a weighted loss function inspired by the work of \cite{schuster2022confident}. 
The purpose is to adapt the output classifier so that it can make early predictions during inference using the output logits from earlier decoder layers (i.e., after different repetitions of decoder layer groups).
More formally, the shared loss function is given by $L_w = \sum_{i=1}^{G} w_{i}L_{i}$, where $G$ is the number of decoder layer groups, and $L_{i}$ and $w_{i}$ correspond to the loss and the applied weighting for group $i$, respectively.
The default weighting scheme we use was the linear weighting described in \cite{schuster2022confident}, which is given as $w_{i}= i / (\sum_{i} i)$. 
Note that this weighting scheme intentionally weights the loss for later layers higher to ensure the final output accuracy is not degraded.
The training scheme using a shared classifier is also illustrated in part (a) of Figure \ref{fig:implementation-overview}.

\subsection{Speculative Pipelined Execution} 
\label{sec:spec_pipe_exec}
Diagrams (b) and (c) in Figure \ref{fig:implementation-overview} outline how the forward pass is performed in our speculative approach.
Our decoding algorithm is outlined in detail in Algorithm \ref{alg:pipelined-decoding} (Appendix \ref{sec:appendix-algorithm}). 
In essence, SPEED speculatively predicts future tokens based on early predictions and then concatenates them with the current token for their parallel processing. 
The crucial feature of SPEED is its \textit{invalidation logic} since speculative predictions can be sometimes incorrect.
To achieve this, our framework keeps track of previous classifications for each token at the previous stage (i.e., before passing through a decoder layer group) and performs the invalidation logic whenever subsequent classifications change after the current stage (i.e., after passing through a decoder layer group). 
In such a case, any future iterations that have been speculatively initiated using the previous classifications must be flushed out and restarted. 

Another crucial implementation detail is that the internal logic in the attention module and the internal Key/Value (KV) cache management logic both need to be modified to facilitate pipelining.
The KV cache corresponds to intermediate activations associated with earlier tokens in the sequence, which are required for calculating later tokens.
The KV cache management logic has to ensure that when future tokens are invalidated, all previous KV cache updates corresponding to these tokens are also invalidated.
These modifications play a key role in ensuring that the final output classification for each token remains unaffected by speculation.

\section{Results}
\label{sec:results}
\begin{figure*}[t!]
  \centering
  \includegraphics[width=\columnwidth]{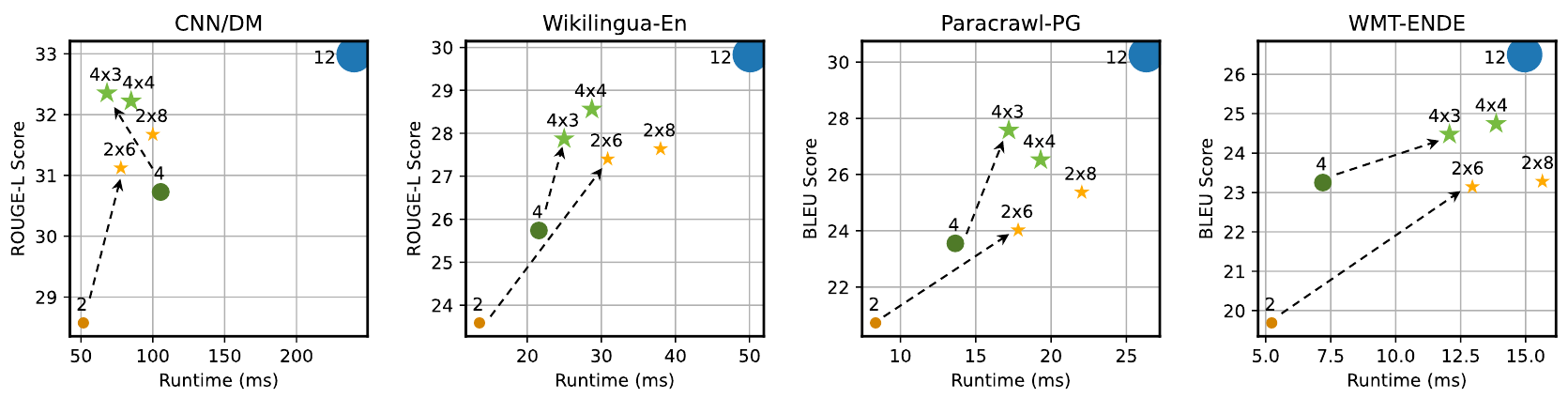}
  \caption{
  Accuracy versus Efficiency for T5-Base with Speculative Execution on an NVIDIA A5000 GPU. 
  The stars correspond to the parameter sharing configurations with speculative pipelined execution, and the circles correspond to the baseline configurations (with either the normal full-length decoder or a shallower decoder).
  The dot colors indicate the models with the same number of parameters (and the dot sizes are proportional to the number of parameters in the decoder). 
  The arrows indicate the accuracy improvements we can get from using parameter sharing without the full latency penalty that we would normally incur. The reported runtime is the average latency across 500 examples in the test set.} 
      \label{fig:pareto-curves-base}
\vspace{-3mm}
\end{figure*}

\subsection{Implementation and Training Details}
\label{sec:appendix-training}

We implement SPEED within the T5X \cite{t5x-framework} repository, which is built on top of the JAX \cite{jax} framework. 
Our implementation for pipelined inference used a custom decoding function in the T5X framework. 
Our initial profiling runs also indicated that the existing greedy decoding function in JAX had greater runtime overhead than our custom decoding algorithm, likely due to additional optional arguments that were unused in our experiments. 
In order to benchmark the networks without parameter sharing, we therefore implemented a stripped-down greedy decoding function to serve as a fair baseline since it has minimal added control logic.

We use the baseline T5-Base decoder-only model architecture \cite{t5}, which has 12 decoder layers, a hidden dimension of 768, 12 attention heads each with dimension 64, and an FFN dimension of 2048 (the default configuration in T5X \cite{t5x-framework}). 
We keep all model architecture parameters constant across all experiments aside from the number of decoder layers. 
We use the 12-layer network as a baseline for comparison since it has the same number of total layers as the configurations with parameter sharing.
We pretrain each network from scratch on C4 \cite{t5}, and we finetune networks on translation and summarization tasks \cite{wmt19translate,cnndailymail,ladhak2020wikilingua,al-ghussin-etal-2023-exploring}. 
For configurations using parameter sharing, parameter sharing is incorporated throughout pretraining and finetuning. 
Additional training details are provided in Appendix \ref{sec:appendix-training}.

\subsection{Main Results}
\label{sec:main}
Figure \ref{fig:pareto-curves-base} shows the accuracy versus efficiency tradeoff comparisons.
When employing parameter sharing with SPEED, we observe significant speedups relative to the baseline 12-layer decoder network, achieving close to the same runtime as the shorter decoder network without parameter sharing across all benchmarks other than WMT-ENDE. 
Additionally, our parameter-sharing configurations attain significantly higher accuracy than the shallow decoder baselines. 
This demonstrates how the SPEED approach allows for improving accuracy for a fixed model size without a significant runtime penalty.
We further experiment with deepening the decoder with parameter sharing by sharing parameters more times such that the total number of layers is increased. 
We find that deepening the decoder generally improves accuracy with minimal runtime penalty; as such, we believe that this is a promising approach to further boost accuracy for a fixed model size without much latency overhead. Appendix \ref{sec:appendix-prediction} provides analysis for the accuracy of predictions made at early layers with \OURS.  A detailed analysis of performance implications of the SPEED approach (and analysis of the lesser speedups we observe for WMT-ENDE) is provided in Appendix \ref{sec:appendix-perf}.
\section{Conclusion}
\label{sec:discussion}

We present a novel decoding strategy that allows for pipelined execution in Transformer decoders with parameter sharing. 
We describe the modifications required to the model architecture to leverage pipelined execution to reduce memory traffic (namely, cyclic parameter sharing in the decoder module).
We observe consistent accuracy gains across all tasks for an equivalent model size, with only a small latency penalty. 
These results demonstrate the accuracy and performance benefits of our pipelined inference approach, showing how \OURS \ allows for deeper decoder configurations with parameter sharing to improve accuracy for a fixed parameter budget and minimal latency penalty.

\bibliography{neurips2023}

\clearpage
\appendix

\section{Related Work}
\label{sec:related-work}

Prior works have explored parameter sharing in encoder-only \cite{lan2020albert}, encoder-decoder \cite{dehghani2019universal}, and decoder-only \cite{reid2021subformer} Transformers as a method for reducing the size of the network by sharing parameters across all layers in the encoder and/or all layers in the decoder. 
\cite{takase2022lessons} explored only sharing parameters amongst a subset of layers and found that cyclic parameter sharing schemes outperformed sharing across all layers.
There have also been several works on speculative decoding which aim to produce a set of ``draft'' tokens autoregressively using a smaller network and then correct them (in parallel) using a larger network  \cite{chen2023accelerating,leviathan2022fast,kim2023big}. 
Our work instead aims to support speculative execution within a single network in order to accelerate inference with parameter sharing networks.

There are also prior works that aim to accelerate decoder inference through early exiting, where inference is terminated early when the model is confident that it can already predict the next token \cite{schuster2022confident, tang2023need}. 
Our work also leverages similar intuition, namely that while certain predictions truly benefit from the models’ full capacity, other continuations are more trivial and can be solved with reduced compute \cite{schuster2022confident}.
However, our proposed approach for accelerating decoder inference has advantages over typical early exiting approaches. 
Although early exit can be applied to an existing network and doesn't require pretraining, our method is guaranteed to always achieve the same accuracy as the baseline network with parameter sharing since it fixes any mistakes. 
Our method also reduces the model size through parameter sharing (in addition to the speedup from pipelined execution).

\section{Algorithm}

\label{sec:appendix-algorithm}

Algorithm \ref{alg:pipelined-decoding} provides the detailed outer-loop decoding algorithm for pipelining decoder inference. The ``iteration\_indices" variable is responsible for both tracking which pipeline stages have a valid token and also what iteration in the sequence these valid tokens are at. 
For example, if the model has 6 groups of shared decoder layers and three valid tokens (corresponding to iterations 3, 2, and 1 in the sequence) which are entering layers 1, 3, and 5 in the network, ``iteration\_indices" will be equal to (3, -1, 2, -1, 1, -1).

\begin{algorithm*}
\caption{Pseudocode for the Pipelined Decoding Algorithm. Note that the encoder outputs are also used by the pipelined forward pass (and the KV cache is updated within this function), but these inputs are omitted for brevity.}
\label{alg:pipelined-decoding}              
\begin{algorithmic}[1]                   
\State \textbf{Constant Values:} 
\State{\textit{max\_decode\_length} is the maximum number of generated tokens}
\State{\textit{G} is the number of groups of shared decoder layers}
\State{\textit{``bos"} is the beginning of sentence token}
\State{\textit{``eos"} is the end of sentence token}
\State{\textit{``pad\_id"} is the padding token}
\State \textbf{Decoding State:} 
\State{\textit{sequence} contains the output sequence (size: max\_decode\_length)}
\State{\textit{previous\_tokens} is an array for keeping track of tokens from the last iteration (size: G - 1)}
\State{\textit{iteration\_indices} is an array for tracking which pipeline stages have valid tokens and what iteration they are at in the sequence (size: G)}
\State{\textit{current\_index} contains the next index to commit a token to in the sequence}
\State{\textit{current\_token} contains the current token to pass in to the model}
\State{\textit{graduated\_token} contains the most recently committed token}
\State \textbf{Temporary Variables:} 
\State{\textit{new\_logits} is the output from the forward pass through a pipeline stage (size: G * number of classes). Note that the number of classes is determined by the tokenizer used (in this work it is fixed at 32K).}
\State{\textit{tokens} is the output classifications from the current forward pass (size: G)}
\State{\textit{compare\_tokens} is a vector that stores comparisons between the current and previous tokens (size: G)}
\State{\textit{last\_invalid} is the index of the oldest token in the pipeline whose classification changed since the last iteration}
\State{\textit{start\_idx} is the index in the sequence of the next token entering the first pipeline stage}

\State \textbf{Functions:} 
\State{\textit{pipeline\_forward\_pass} computes the forward inference pass for a single stage}
\State{\textit{get\_index\_of\_last\_nonzero} gets the index of the last nonzero in the input array, and otherwise returns 0}
\State{\textit{range(N)} generates the array [0,1,...,N-1]}
\State{\textit{argmax(A)} returns an array containing the index of the maximum value for each column in A}
\algstore{myalg}
\end{algorithmic}
\end{algorithm*}

\begin{algorithm*}             
\begin{algorithmic}[1]                   
\algrestore{myalg}

\State \textbf{Initialize:} 
\State sequence = [``pad\_id" for i in range(max\_decode\_length)] 
\State previous\_tokens = [-1 for i in range(G-1)] 
\State iteration\_indices = [-1 for i in range(G)] 
\State current\_index = 0 
\State current\_token = ``bos"
\State graduated\_token = -1 

\While{((graduated\_token != ``eos") and current\_index \textless \space max\_decode\_length)} 
  \State new\_logits = pipeline\_forward\_pass(current\_token, iteration\_indices)
  \State tokens = argmax(new\_logits) 
  \State tokens[iteration\_indices == -1] = -1 
  \State compare\_tokens = [0, tokens[1:] != previous\_tokens] 
  \State last\_invalid = get\_index\_of\_last\_nonzero(compare\_tokens) 

  \State iteration\_indices[i \textless \space last\_invalid] = -1 
  \State tokens[iteration\_indices == -1] = -1 
  \State previous\_tokens = tokens[:-1] 
  \State current\_token = tokens[last\_invalid] 
  \State start\_idx = iteration\_indices[last\_invalid] + 1

  \If{(iteration\_indices[-1] != -1)}
    \State sequence[current\_index] = tokens[-1] 
    \State graduated\_token = tokens[-1] 
    \State current\_index += 1 
  \EndIf

  \If{(start\_idx \textless \space max\_decode\_length)}
        \State iteration\_indices = [start\_idx, iteration\_indices[:-1]]
  \Else
        \State  iteration\_indices = [-1, iteration\_indices[:-1]]
  \EndIf
\EndWhile
\end{algorithmic}
\end{algorithm*}

\clearpage

\section{Training Details}

\label{sec:appendix-training}

We used the default SentencePiece tokenizer with a vocabulary size of 32K, and we used tied input and output embeddings \cite{sentencepiece}.
We pretrained on C4 (Colossal Clean Crawled Corpus) for 524,288 steps using a batch size of 128 \cite{t5}. 
C4 is a large dataset of filtered English text scraped from the web \cite{t5}.
We used a base learning rate of 1 with a square root decaying learning rate scheduler and with 10K warmup steps.
We focused on two particular sequence-to-sequence tasks during finetuning: translation and summarization. 
For translation, we used the WMT English to German dataset as well as the English to German Paracrawl-Paragraph translation dataset \cite{wmt19translate, al-ghussin-etal-2023-exploring}. The Paracrawl-Paragraph dataset was used to also evaluate on a translation dataset with longer source and target context lengths (since it consists of full paragraph translations).
For summarization, we used the CNN/DailyMail dataset, which consists of news articles written by journalists at CNN and the Daily Mail \cite{cnndailymail}, as well as the English to English split of the Wikilingua multilingual summarization dataset \cite{ladhak2020wikilingua}.
We finetuned for 262,144 steps using a batch size of 128 and dropout of 0.1, using input/target sequence lengths of 512/512 across all tasks. 
When finetuning, we used a constant learning rate of 0.001 with 1K warmup steps.
We evaluated checkpoints every 5,000 steps during finetuning on the validation set, and then reported results on the test set using the checkpoint with the best accuracy on the validation set.
Both training and inference arithmetic were performed in BF16 precision.
We used TPU v2-8 machines on Google Cloud Platform for training experiments, and we launched these experiments using Skypilot \cite{skypilot}.

\section{Performance Analysis}
\label{sec:appendix-perf}

\subsection{Prediction Consistency}

\label{sec:appendix-prediction}

\begin{figure*}[t!]
    \centering
    \includegraphics[width=\columnwidth]{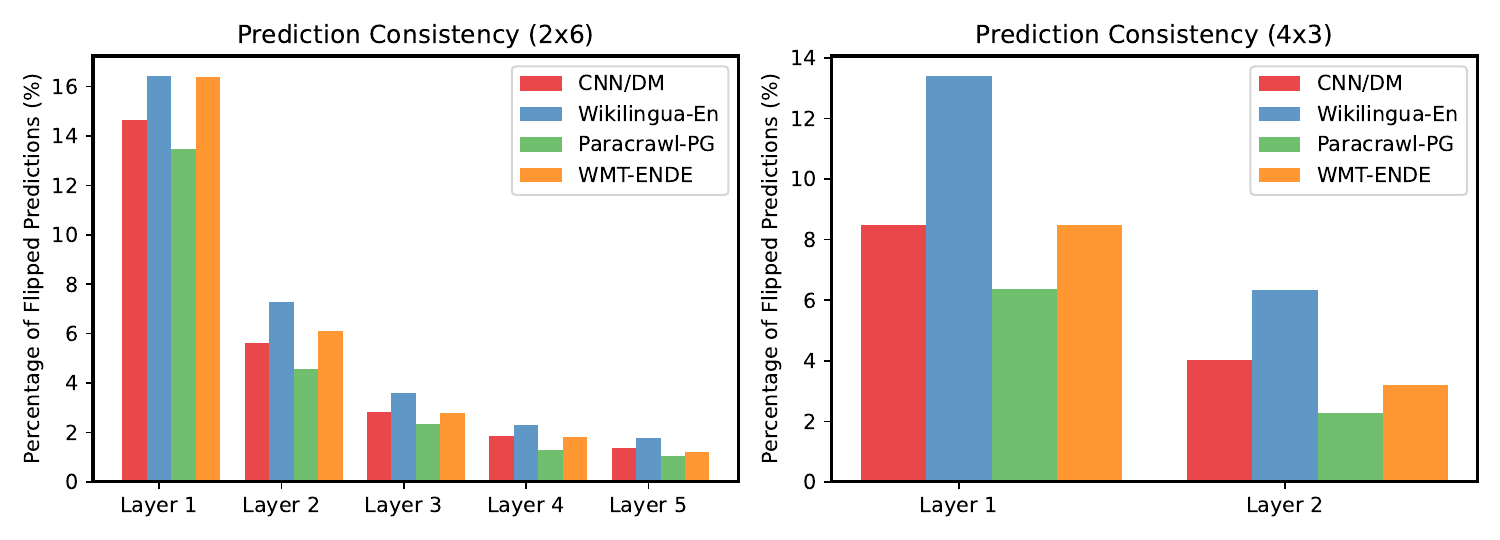}
    \caption{Proportion of predictions that were flipped at each layer when using speculative pipelined execution with 2x6 and 4x3 configurations.}
    \label{fig:Prediction-consistency-plots}
\end{figure*}
In order to assess the accuracy of the predictions made from our network at earlier layers, we profiled the proportion of predictions that were flipped between pairs of layers during inference. 
Figure \ref{fig:Prediction-consistency-plots} shows the prediction consistencies for 4x3 and 2x6 network configurations across all tasks. 
Upon examining these numbers and plots, we found that the model is able to make the majority of predictions accurately at early layers. Across all three configurations, the proportion of predictions that would need to be corrected after the first layer was between 
13-17\% for the 2x6 configuration and between 6-14\% for the 4x3 configuration, showing that the majority of predictions were correct at early layers. 
Additionally, we found that a very small percentage of predictions flipped at later layers. 
This shows that the model tends to converge to the final answer and does not experience much oscillation between different predictions.

\subsection{WMT-ENDE Performance}
\label{sec:appendix-perf-wmtende}

The primary reason that we observed greater latency penalties with our approach for WMT-ENDE compared with the other translation and summarization tasks was due to its shorter output generation lengths.
The benefits from our pipelined decoding approach come from being able to process multiple tokens in parallel, and in the first few and last few iterations with our method, there will be fewer tokens in the pipeline.
This means that the first iterations and final iterations in pipelined decoding aren't completely overlapped.
This is only a limiting factor for tasks with shorter generated sequence lengths (where the average number of tokens generated is close to the number of shared groups of layers in the network). 
The generation lengths for WMT-ENDE are typically shorter than summarization tasks and paragraph-level translation, which leads to increased latency penalties.

\subsection{CNN/DM Performance}
With CNN/DM, we actually observed reduced latency when inferring the 4x3/4x4 configurations relative to the 4-layer network without parameter sharing. 
This is unexpected, since even assuming perfect prediction for the networks with parameter sharing, the latency would not be less than the baseline 4-layer network (assuming the same output generation length).
However, it is possible for the parameter sharing configurations to exhibit lower latency due to differences in the average generation lengths for the networks with parameter sharing relative to the network without parameter sharing (as if the average generation length is shorter for the networks with parameter sharing, they could have lower average latency). 

\subsection{General Discussion}
There are several factors which impact the runtime when employing SPEED. 
\begin{itemize}
    \item One factor is the generation length, as the first iterations and final iterations in pipelined decoding aren't completely overlapped (as discussed in Appendix \ref{sec:appendix-perf-wmtende}).
    This limits the runtime gains from \OURS \ for tasks with short output generation lengths.
    \item Because the embedding matrix is large, it can actually end up consuming a large portion of the memory bandwidth (and hence the runtime) for smaller models.
    This is a crucial reason why the latency gains aren't linear as you go from a 12-layer network down to a 2-layer network even without considering parameter sharing or speculative execution.
    \item One additional performance implication is that if the pipeline is too deep (i.e. layers are shared too many times), this can lead to greater misprediction penalties.
    Improving prediction consistency is therefore crucial for improving runtime with deeper decoder configurations. 
\end{itemize}


\end{document}